\documentclass[runningheads]{llncs}
\usepackage{graphicx}
\usepackage{amsmath,amssymb} % define this before the line numbering.
\usepackage{color}
\usepackage{algorithm,algorithmic}
\usepackage[misc]{ifsym}
\usepackage{hyperref}
\makeatletter
\renewcommand*{\@fnsymbol}[1]{\ensuremath{\ifcase#1\or \dagger\or \star\or \ddagger\or
    \mathsection\or \mathparagraph\or \|\or **\or \dagger\dagger
    \or \ddagger\ddagger \else\@ctrerr\fi}}
\makeatother

\begin{document}
\title{Discriminative Region Proposal Adversarial Networks for High-Quality Image-to-Image Translation} 
% Replace with your title

\titlerunning{Discriminative Region Proposal Adversarial Networks (DRPAN)}
% Replace with a meaningful short version of your title
%
\author{Chao Wang\orcidID{0000-0002-1177-8104} \and
Haiyong Zheng*\orcidID{0000-0002-8027-0734} \and
Zhibin Yu\orcidID{0000-0003-4372-1767} \and
Ziqiang Zheng\orcidID{0000-0003-4027-7955} \and
Zhaorui Gu\orcidID{0000-0002-6673-7932} \and
Bing Zheng\orcidID{0000-0003-2295-3569}
}
%
%Please write out author names in full in the paper, i.e. full given and family names. 
%If any authors have names that can be parsed into FirstName LastName in multiple ways, please include the correct parsing, in a comment to the volume editors:
%\index{Lastnames, Firstnames}
%(Do not uncomment it, because you may introduce extra index items if you do that, we will use scripts for introducing index entries...)
\authorrunning{C. Wang, H. Zheng, Z. Yu, Z. Zheng, Z. Gu and B. Zheng}
% Replace with shorter version of the author list. If there are more authors than fits a line, please use A. Author et al.
%
\institute{Ocean University of China, Qingdao 266100, China\\
\email{chaowangplus@gmail.com},
\email{\{zhenghaiyong, yuzhibin\}@ouc.edu.cn},\\
\email{zhengziqiang@stu.ouc.edu.cn},
\email{\{guzhaorui, bingzh\}@ouc.edu.cn}\\
\url{http://vision.ouc.edu.cn}\\
* Corresponding author
}
\maketitle              % typeset the header of the contribution
\begin{abstract}
Image-to-image translation has been made much progress with embracing Generative Adversarial Networks (GANs). However, it's still very challenging for translation tasks that require high quality, especially at high-resolution and photorealism. In this paper, we present Discriminative Region Proposal Adversarial Networks (DRPAN) for high-quality image-to-image translation. We decompose the procedure of image-to-image translation task into three iterated steps, first is to generate an image with global structure but some local artifacts (via GAN), second is using our DRPnet to propose the most fake region from the generated image, and third is to implement ``image inpainting'' on the most fake region for more realistic result through a reviser, so that the system (DRPAN) can be gradually optimized to synthesize images with more attention on the most artifact local part. Experiments on a variety of image-to-image translation tasks and datasets validate that our method outperforms state-of-the-arts for producing high-quality translation results in terms of both human perceptual studies and automatic quantitative measures.

\end{abstract}
\section{Introduction}
From the aspect of human visual perception, why we consider a synthesized image as fake is often because it contains local artifacts. Although it looks like real at the first glance, we can still easily distinguish the fake from the real by gazing for only about $1000ms$~\cite{Chen2017Photographic}. Human being has the ability to draw a realistic scene from coarse structure to fine detail, that is, we usually get the global structure of a scene while focus on the detail of an object and understand how it is associated with surroundings. Under this intuition, our goal of this work is to develop an image-to-image translation system for high-quality image synthesis with clear structure and vivid details.

Many efforts have been made to develop an automatic image-to-image translation system. The straightforward approach was to optimize on pixel-wise space with L1 or L2 loss~\cite{Dong2016Image,Long2015Fully}. However, both of them suffer from blur problem. So some works added adversarial loss for generating more sharp images in both spatial and spectral dimensions~\cite{Isola2016Image}. Except for the GAN loss, perceptual loss has been used in image-to-image translation tasks, but it was limited to a pre-training deep model and the training datasets~\cite{Wang2017Perceptual}. Although we have a variety of losses to evaluate the discrepancy between real image and generated image, using GAN for image-to-image translation still encounters with the artifacts and unsmooth color distribution problems, and it is even hard to generate high-resolution photo-realistic images because of the high dimension distribution~\cite{Odena2016Conditional}.

\begin{figure}[!ht]
  \centering
  \includegraphics[width=\textwidth]{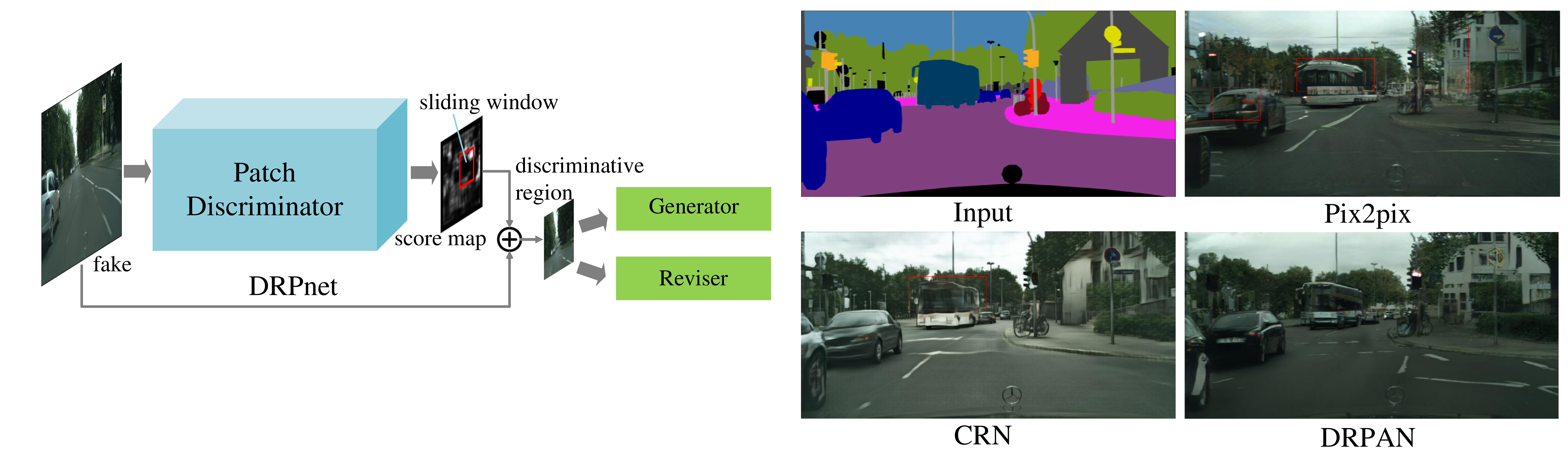}
  \caption{\textbf{Left}: Our Discriminative Region Proposal network (DRPnet). \textbf{Right}: Synthesized samples compared with previous works on Cityscapes validation dataset~\cite{Cordts2016The}. The regions within red window show obvious artifacts or deformation. Our method can synthesize images with clear structure and vivid details.}
  \label{fig:DRPnet}
\end{figure}

So, how could we solve this problem intuitively? We decompose the procedure of image-to-image translation task into three iterated steps, first is to generate an image with global structure but some local artifacts (via GAN), second is to propose the most fake region from the generated image (using our DRPnet shown in Fig.~\ref{fig:DRPnet}), and third is to implement ``image inpainting'' on the most fake region for more realistic result, so that the system (our DRPAN) can be gradually optimized to synthesize images with more attention on the most artifact local part. Inspired by this motivation, we develop a framework based on patch-wise discriminator to predict the discriminative score map and use sliding windows to find the most artificial region. Then the proposed discriminative region will be used to mask the corresponding real sample and output as ``masked fake''. Finally, we propose a reviser to distinguish the real from the masked fake for producing realistic details and serve as auxiliaries for generator to synthesize high-quality translation results. The reviser will critic on the fake image iteratively with different regions. We provide a weighted parameter to balance the contribution of the patch discriminator and our reviser for different levels of translation tasks. Using this proposed DRPAN, we can synthesize high-quality images with high-resolution and photo-reality details but less artifacts.

The main contribution of the study is threefold: first, we design the mechanism to explore patch-based discriminators for producing discriminative region; second, we propose the reviser for GANs to provide constructive revisions for generator which usually are missed by patch discriminator; third, we build a DRPAN model as a general-purpose solution for high-quality image-to-image translation tasks on different levels. The code of this paper is available at \url{https://github.com/godisboy/DRPAN}.

\section{Related works}
\textbf{Feed-forward based approach.} Deep Convolutional Neural Networks (CNNs) have been performed well on many computer vision tasks. For style transform problems~\cite{Johnson2016Perceptual}, many studies were mainly based on VGG-16 network architecture~\cite{LiW16b} and used perceptual losses for style translation~\cite{gatys2015neural}. Network architectures that work well on object recognition tasks have been proved to work well on generative models, \emph{e.g.}, some computer vision translation and editing tasks used residual block as a strong feature learning representation architecture~\cite{Ledig2016Photo,Lin2016RefineNet}. Feed-forward CNNs accompanied with per-pixel loss have been presented for image super-resolution~\cite{Dong2016Image,Kim2016Accurate,Shi2016Real,Johnson2016Perceptual}, image colorization~\cite{Deshpande2015Learning,Zhang2016Colorful}, and semantic segmentation~\cite{Long2015Fully,chen2015semantic,ronneberger2015unet}. A recent work for photo realistic image synthesis system, called CRN~\cite{Chen2017Photographic}, can synthesize images with high resolution. However, the images synthesized by feed-forward based approach usually become smooth too much rather than realistic, \emph{i.e.}, not sharp enough in details. Besides, these methods are limited to be applied to other image-to-image translation tasks.

\textbf{GAN based approach.} GANs~\cite{Goodfellow2014Generative} introduced an unsupervised method to learn real data distribution. And DCGAN~\cite{RadfordMC15} firstly used CNNs to train generative adversarial networks which was hard to be deployed in other tasks before. Then, CNNs were extensively used for designing GAN architectures. Towards stable training of GAN, WGAN~\cite{Arjovsky2017Wasserstein} replaced Jensen-Shannon divergence by Wasserstein distance as the optimization metric, and recently a variety of more stable alternatives have been proposed~\cite{Qi17,KodaliAHK17,Gulrajani2017Improved}. Wang and Gupta~\cite{Wang2016Generative} combined structured GAN with style GAN to learn to generate natural indoor scenes. Reed \emph{et al.}~\cite{Reed2016Generative} used text as conditional input to synthesize images with semantic variation. Pathak \emph{et al.}~\cite{Pathak2016Context} proposed context encoders for image inpainting accompanied by adversarial loss. Li \emph{et al.}~\cite{Li2017Generative} trained GANs with a combination of reconstruction loss, two adversarial losses and a semantic parsing loss for face completion. Nguyen \emph{et al.}~\cite{Nguyen2016Plug} presented Plus and Play Generative Networks for high-resolution and photo-realistic image generation with the resolution of $227 \times 227$ images. Isola \emph{et al.} explored~\cite{Isola2016Image} conditional GANs for a variety of image-to-image translation problems. ID-CGAN~\cite{Zhang2017Image} combined conditional GANs with perceptual loss for single image de-raining and de-snowing. Considering that the paired images are less and hard to collect, some works proposed unpaired or unsupervised translation frameworks~\cite{ZhuPIE17,Kim2017Learning,YiZTG17}. But it limits to the similarity of translation between source domain $A$ and target domain $B$.

PatchGAN was firstly used in neural style transfer with CNNs based on patch feature inputs~\cite{LiW16b}. Pix2pix~\cite{Isola2016Image} showed that a full ImageGAN does not show quality improvement compared with a low $70 \times 70$ patch discriminator which has less parameters and needs low computing resource. SimGAN~\cite{Shrivastava2016Learning} used patch based score map for real image synthesis tasks and mapped a full image to a probability map. Our method explores PatchGAN to a unified discriminative region proposal network model for deciding where and how to synthesize via a reviser. We show that this approach can improve translation results on high-quality, especially at high-resolution and photo-reality.

\section{Method}
\begin{figure}[!ht]
  \centering
  \includegraphics[width=\textwidth]{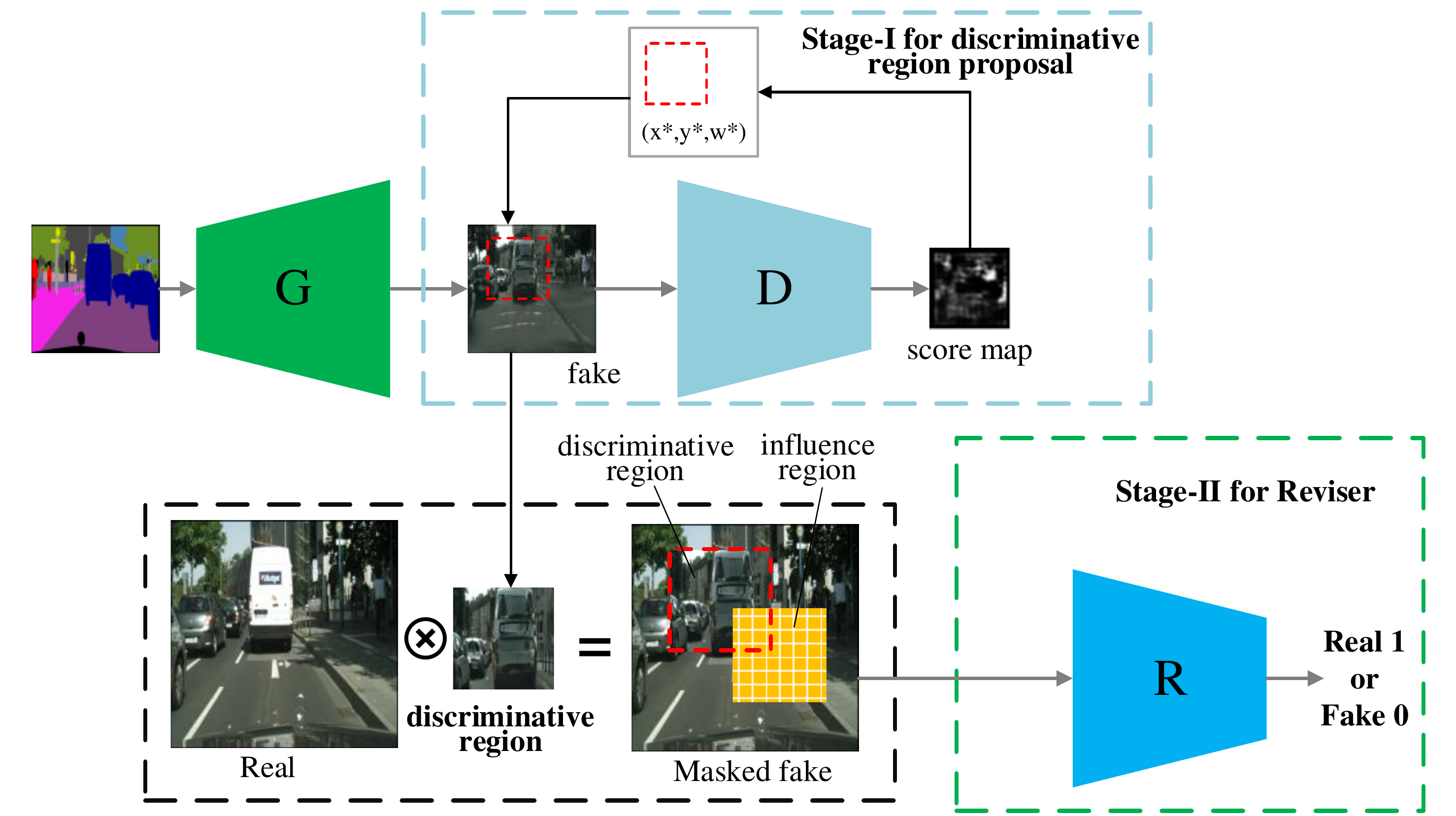}
  \caption{The overall network architecture and data flow of our proposed Discriminative Region Proposal Adversarial Networks (DRPAN), which is composed of three components: a generator, a discriminator, and a reviser, and is a unified model for image-to-image translation tasks.}
  \label{fig:DRPAN}
\end{figure}

Our image-to-image translation model, called Discriminative Region Proposal Adversarial Networks (DRPAN), is composed of three components: a generator, a discriminator, and a reviser. The discriminator explores PatchGAN to construct Discriminative Region Proposal network (DRPnet, see Fig.~\ref{fig:DRPnet}) to find and extract the discriminative region for producing masked fake sample, while the reviser adopts CNN to distinguish the real from the masked fake to provide constructive revisions for generator. The overall network architecture and data flow are illustrated in Fig.~\ref{fig:DRPAN}. 

\begin{figure}[!ht]
  \centering
  \includegraphics[width=\textwidth]{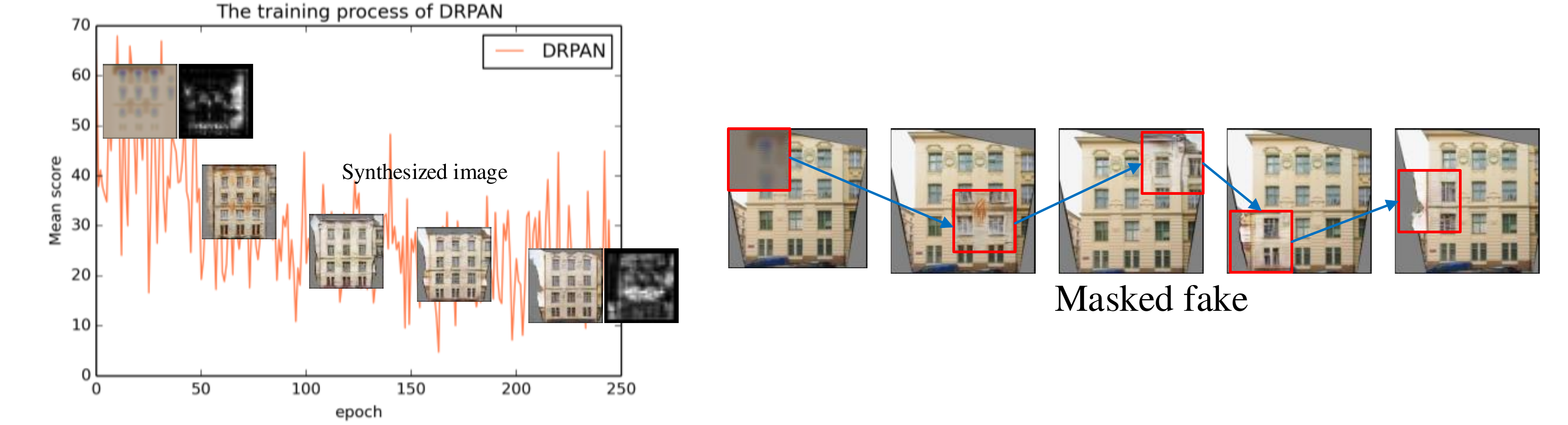}
  \caption{The training process of DRPAN on facades dataset~\cite{tylevcek2013spatial}. \textbf{Left}: The plotting curve shows mean value of score map on synthesized samples. \textbf{Right}: Step by step synthesis on different discriminative regions.}
  \label{fig:processing}
\end{figure}

Fig.~\ref{fig:processing} shows our process of how to improve the quality of synthesized image. It can be seen that, as our DRPAN continues to train, the discriminative region for masked fake images (right) varies so that the quality of synthesized images (left) are improved with brighter score map (the first and the last). Besides, although it is hard to distinguish the synthesized sample from the real sample after many epochs, our DRPAN can still revise the generator to optimize the synthesized result in the details for high quality.

\subsection{DRPAN}
\begin{figure}[!ht]
  \centering
  \includegraphics[width=.8\textwidth]{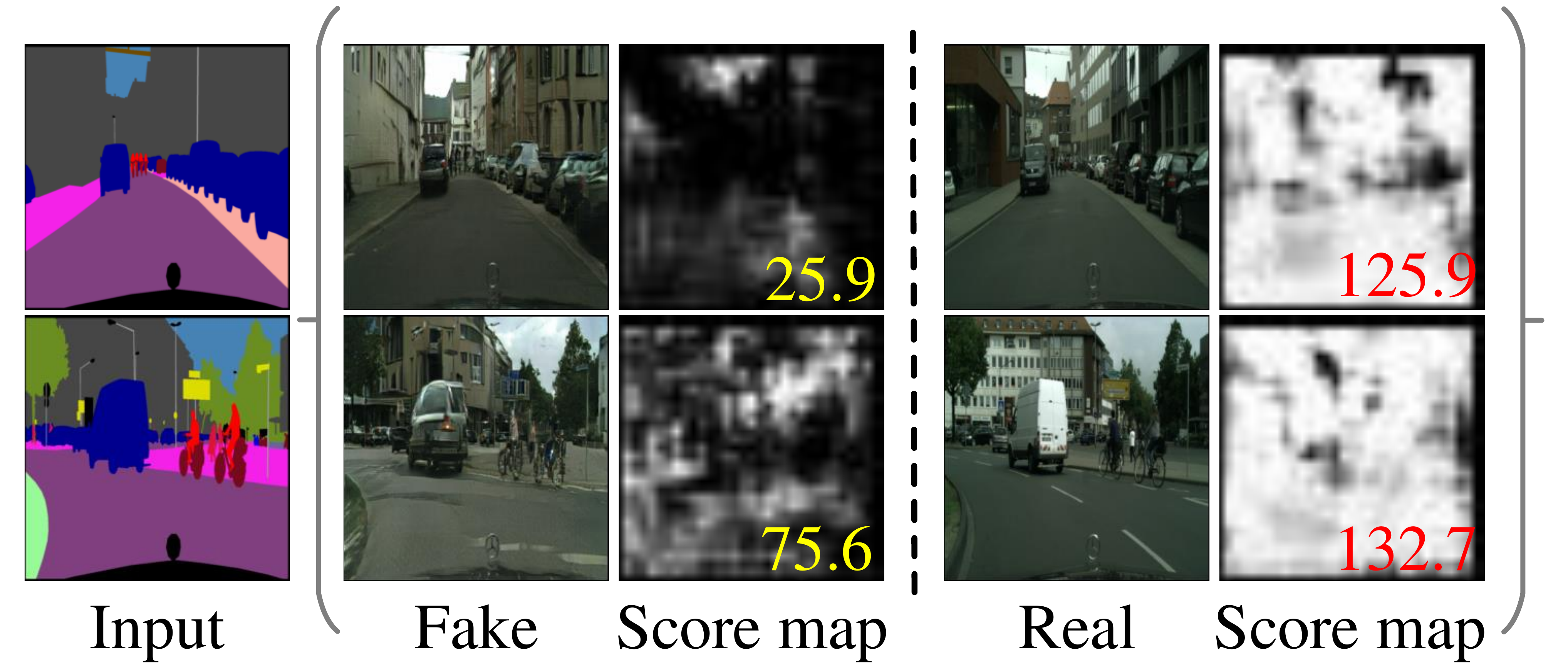}
  \caption{The output results of score map on different quality levels (fake and real) of images by a pre-trained PatchGAN. The darkest regions on score maps mean the lowest quality, indicating that patch-based discriminators can be explored for discriminative region proposal.}
  \label{fig:scoremap}
\end{figure}

We first suggest that patch-based discriminators produce meaningful score maps, which may have applications beyond image synthesis. Fig.~\ref{fig:scoremap} shows the output results of score map on different quality levels (fake and real) of images by a pre-trained PatchGAN. It can be seen that, the score maps of the fake samples, which have obvious artifacts and shape deformation on some regions, are almost dark with lower score on the corresponding regions; in contrast, the score maps of the real samples are brightest with the highest scores. From the visualization of score maps, we can find the darkest region for proposing the discriminative region that indicates the remarkable fake region.

Based on the observation shown in Fig.~\ref{fig:scoremap}, we explore patch discriminator to DRPnet for producing discriminative region. Given an input image with resolution $w_i\times w_i$, and it is processed by the patch discriminator to be a probability score map with size $w_s\times w_s$. Suppose we want to obtain the discriminative region at $w^*\times w^*$, the size of sliding window $w$ for score map can be calculated by
\begin{equation}
w=w^*\times w_s/w_i.
\end{equation}
Then our DRPnet will find the discriminative square patch on score map with the center coordinates $(x_c,y_c)$ and length $w$, so the scale $\tau$ between the input image and output score map is
\begin{equation}
\tau=\frac{w_i-w*}{w_s-w}.
\end{equation}
The center coordinates $(x_c^*,y_c^*)$ of discriminative region will be calculated by
\begin{equation}
\begin{cases}
x_c^*=\tau\times x_c,\\
y_c^*=\tau\times y_c.
\end{cases}
\end{equation}
Finally, the discriminative region $d_r$ produced by DRPnet can be expressed as
\begin{equation}\label{eq:dr}
d_r=F_{\text{DRPnet}}(x_c^*,y_c^*,w^*).
\end{equation}

Instead of only optimizing the independent local regions, we consider the relationship between fake discriminative region and real surrounding influence regions, so that it can connect the fake to the real for providing constructive revisions to generator. The influence region is defined as the region which is connected to the “most fake regions” and has semantic and spatial relationship with the content in it (e.g., the wheel is often below the car window). For this purpose, we mask the corresponding real sample using the fake discriminative region to make masked fake sample, and then design a reviser using CNN to distinguish real from masked fake to optimize the generator for synthesizing high-quality images. The reviser we proposed can also be used for other GANs to improve the quality of generated samples.

\subsection{Objective}
For image-to-image translation tasks, we not only want to generate the realistic samples, but also desire diversity with different conditional inputs. The original GANs suffer from unstability and mode collapse problems~\cite{Arjovsky2017Wasserstein,arora2017generalization}. So some recent works~\cite{Arjovsky2017Wasserstein,Qi17,Gulrajani2017Improved} improved the training of GAN. To stably train our DRPAN with high-diversity synthesis ability, we modify DRAGAN~\cite{KodaliAHK17} as the loss of our reviser $R$, and use the original objective function for training Patch Discriminator.
\begin{equation}\label{eq:d}
%\begin{split}
\mathcal{L}_{D}(G, D_{p}) = \mathbb{E}_{y}[\log D_{p}(x,y)]
 + \mathbb{E}_{x,z}[\log (1-D_{p}(x,G(x,z)))].
%\end{split}
\end{equation}

For reviser $R$, to distinguish between the very similar real and masked fake $y_{\text{mask}}=M(G(x,z))$ ($M(\cdot)$ represents the mask operation), we add a regularization to the loss of reviser as the penalty, which is expressed as
\begin{equation}\label{eq:r}
%\begin{split}
\mathcal{L}_{R}(G, R) = \mathbb{E}_{y}[\log R(x,y)] + \mathbb{E}_{x,z}[\log (1-R(x,y_{\text{mask}}))]
 + \alpha\mathbb{E}_{x, \delta}[\left \|\nabla_{x}R(x+\delta)\right \|-1]^{2},
%\end{split}
\end{equation}
where $\alpha$ is hyper parameter, $\delta$ is random noise on $x$, and $\nabla$ indicates gradient.

Previous studies have found it beneficial to mix the GAN objective with a more traditional loss, such as L2 and L1 distance~\cite{Isola2016Image,Shrivastava2016Learning}. Considering that L1 distance encourages less blurring than L2~\cite{Isola2016Image}, we provide extra L1 loss for regularization on the whole input image and the local discriminative region to generator, which is defined as
\begin{equation}\label{eq:l1}
%\begin{split}
\mathcal{L}_{L_1}(G) = \beta\mathbb{E}_{x,y,z}[\left \| y-G(x,z) \right \|_{1}]
 + \gamma\mathbb{E}_{d_r,y_r,z}[\left \| y_{r}-F_{\text{DRPnet}}(G(x,z)) \right \|_{1}],
%\end{split}
\end{equation}
where $\beta$ and $\gamma$ are hyper parameters, $d_r$ is the discriminative region, and $y_r$ represents the region on the real image corresponding to the discriminative region on the synthesized image.
Then the total loss of generator can be expressed as
\begin{equation}\label{eq:g}
%\begin{split}
\mathcal{L}_{G}(G, D_{p}, R) = -\mathbb{E}_{x,z}[\log (1-D_{p}(x,G(x,z)))]
 - \mathbb{E}_{x,z}[\log (1-R(x,y_{\text{mask}}))] + \mathcal{L}_{L_1}(G).
%\end{split}
\end{equation}

Our proposed model totally contains a generator $G$, a patch discriminator $D_p$ for DRPnet, and a reviser $R$. $G$ will be optimized by $D_{p}$, $R$ and $L_1$. And our full objective function is
\begin{equation}\label{eq:drpan}
L(G, D_{p}, R) = (1-\lambda)\mathcal{L}_{D}(G, D_{p}) + \lambda \mathcal{L}_{R}(G, R) + \mathcal{L}_{L_1}(G),\\
\end{equation}
where $\lambda$ is a hyper parameter to balance $\mathcal{L}_D$ and $\mathcal{L}_R$.

\subsection{Network architecture}
For our generator, we use architecture based on~\cite{Ledig2016Photo} which has convincing power for single image super-resolution. We adopt convolution and fractionally convolution blocks for down and up sampling respectively, and $9$ residual blocks~\cite{ZhuPIE17} for task learning. Each layer uses Batch Normalization~\cite{Ioffe2015Batch} and ReLU~\cite{nair2010rectified} as activation function. For patch discriminator, we mainly implement with $70\times 70$ PatchGAN~\cite{LiW16b,Isola2016Image}. The DRPAN reviser is a discriminator modified on DCGAN~\cite{RadfordMC15} that has a global view on the whole input. At the end of both discriminator and reviser, we adopt Sigmoid as activation function to output probability.% We also deploy the model with more residual blocks and discriminative layers for high-quality image-to-image translation tasks.

\section{Experiments}
\label{sec:experiments}
To evaluate the performance of our proposed method on image-to-image translation tasks, we deploy a variety of experiments about different levels of translation tasks to compare our method with state-of-the-arts. And for different tasks, we also use different evaluation metrics including human perceptual studies and automatic quantitative measures.

\subsection{Evaluation metrics}
\textbf{Image quality evaluation.} PSNR, SSIM~\cite{Wang2004Image} and VIF~\cite{Sheikh2006Image} are some of the most popular evaluation metrics in low-level computer vision tasks such as deblurring, dehazing and image restoration. So for de-raining and aerial to maps tasks, we adopt PSNR, SSIM, VIF and RECO~\cite{baroncini2009polar} to qualify the performance of results.

\textbf{Image segmentation evaluation metrics.} We use standard metrics from Cityscapes benchmark~\cite{Cordts2016The} to evaluate real to semantic labels task on Cityscapes dataset, including per-pixel accuracy, per-class accuracy, and Class IOU.

\textbf{Amazon Mechanical Turk (AMT).} AMT~\cite{Isola2016Image,ZhuPIE17,YiZTG17} is adopted in many tasks as a gold metric to evaluate how real the synthesized images, and we use it as evaluation metric for semantic labels to photo and maps to aerial tasks.

\textbf{FCN-8s score.} 
The intuition of using an off-the-shelf classifiers for automatic quantitative measurement is that if the generated images are realistic, classifiers trained on real images will be able to classify the synthesized image correctly as well~\cite{Isola2016Image}. We use the FCN-8s score~\cite{Long2015Fully} to evaluate semantic labels to real task on Cityscapes dataset. The FCN-8s model trained on Cityscapes segmentation tasks is taken from~\cite{Isola2016Image}.

\subsection{Why DRPAN?}
To study the influence of DRPAN for revising synthesis and different situations of loss between proposed region and real region. We set an experiments which start from a pre-trained PatchGAN and continue for several training pipelines: continue training with PatchGAN; continue training with PatchGAN and L1 loss of discirminative and real region; continue training with PatchGAN and reviser.

\begin{figure}[!ht]
  \centering
  \includegraphics[width=\textwidth]{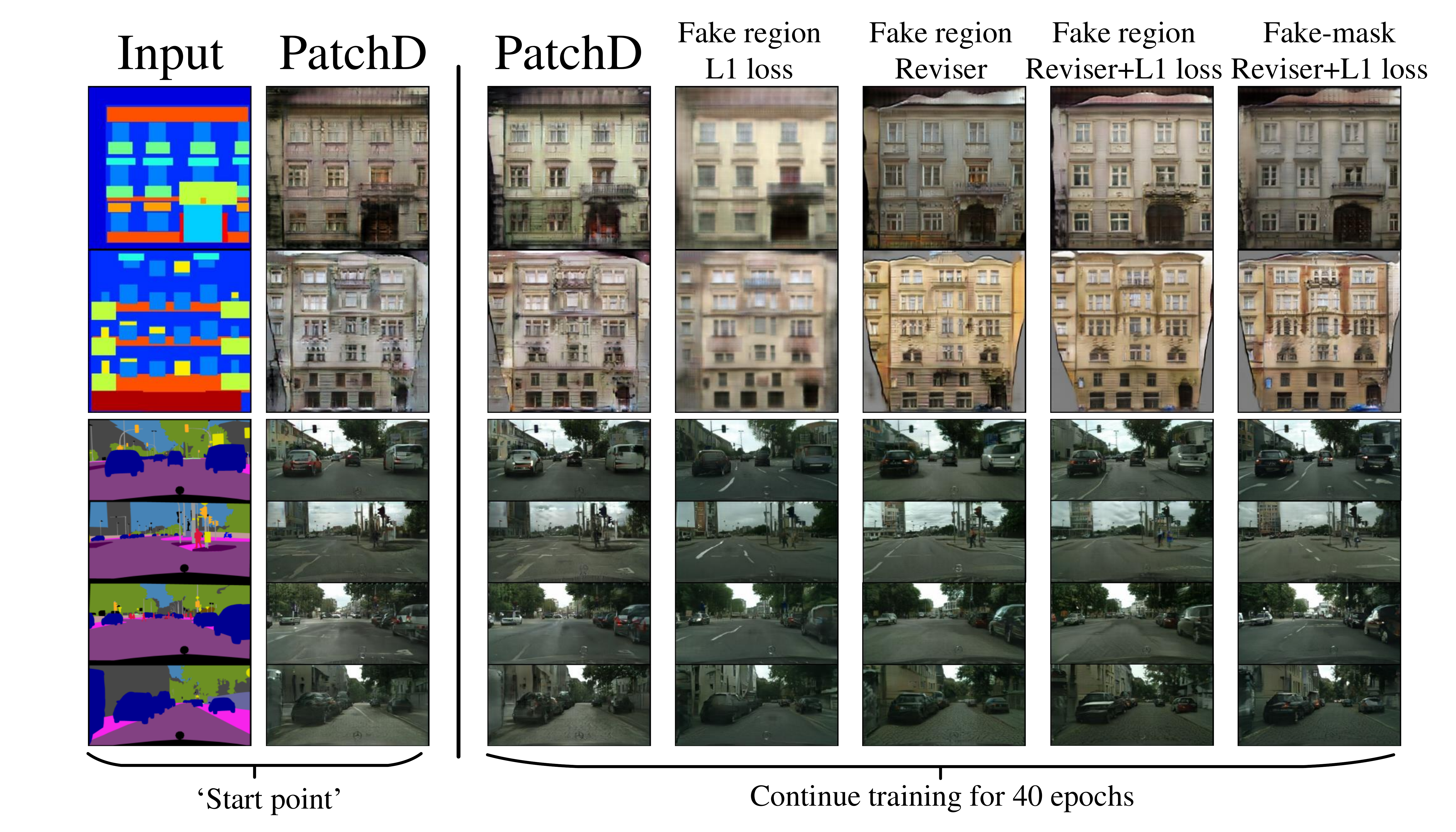}
  \caption{Different methods with various losses produce different quality of results. The second column is the start point of comparison trained by PatchD, and all other models are continued trained for 40 epochs more. These experiments validate the necessary of our DRPnet for discriminative region proposal, our reviser for optimizing generator, and our fake-mask operation for improving synthesis.}
  \label{fig:analysis}
\end{figure}

We argue that the PatchD is efficient to discover the most fake or real region (Fig~\ref{fig:scoremap}) from the image but is limited to improve these regions with fine details for that PatchD is hard to capture the high dimension distribution. In this case, we propose a DRPnet (explore the strength of PatchD) for discriminative region proposal and design a reviser to gradually remove visual artifacts, and thus reduce it to lower dimension estimation problem. This can be seen as a ``top-down'' procedure which is different from other gradually ``bottom-up'' image generation method~\cite{ZhangXLZHWM16}. Fig~\ref{fig:analysis} shows the necessity of our proposed DRPAN for high-quality image-to-image translation, which illustrates that continue training PatchD is no help to reduce artifacts even with a L1 loss for balance, and DRPAN with only L1 loss can smooth the artifacts but not very sharp in details, while DRPAN with reviser exceeds the PatchD's performance with less visual artifacts. The combination of reviser and L1 loss can reduce these artifacts ignored by PatchD. We also find that fake-mask operation can improve the fluency of whole image in certain samples (\emph{e.g.}, the connection between door and wall). So DRPAN with fake-mask is implemented in the following experiments.

% analysis of different loss

\subsection{Low level translation}
We first apply our model on two low level translation tasks which are only related to the appearance translation of images, for example, in de-raining task we don't need change the content and texture of the input sample. So we set $\lambda = 1$ in Eqn.~\ref{eq:drpan} for image synthesis using only reviser.

\begin{figure}[!ht]
  \centering
  \includegraphics[width=.9\textwidth]{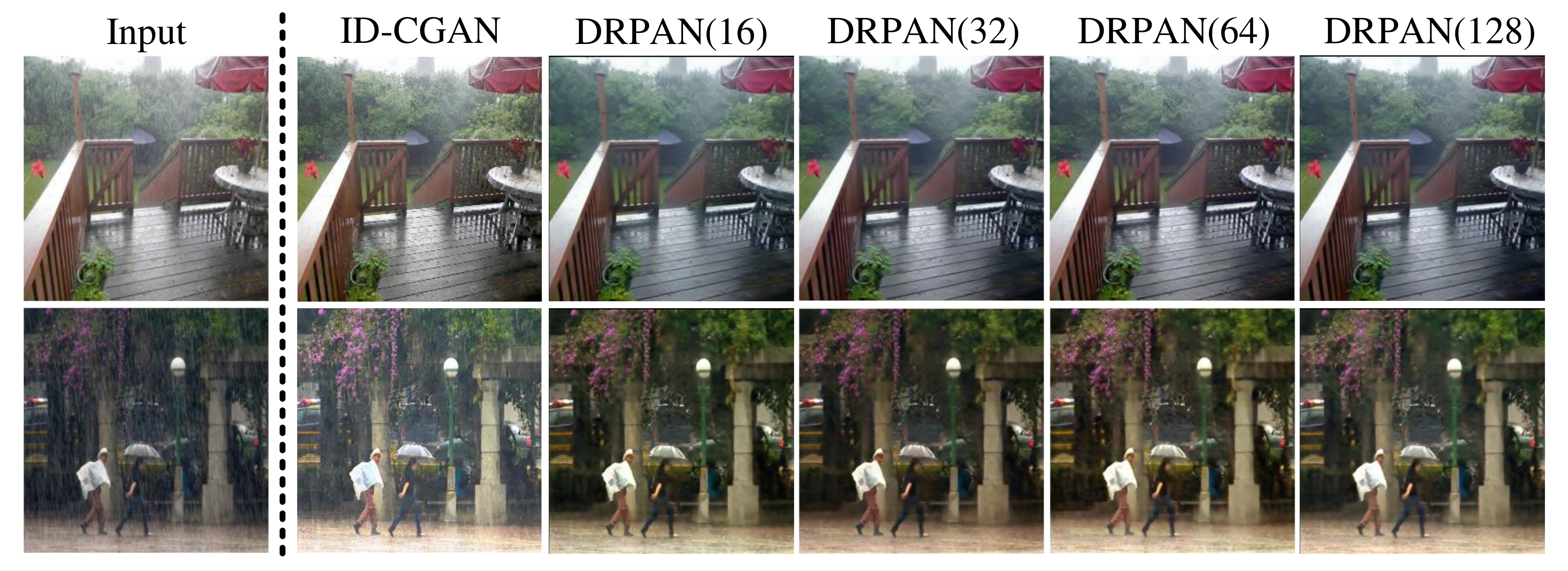}
  \caption{Example results of our DRPAN with different sizes of discriminative region compared to ID-CGAN~\cite{Zhang2017Image} on single image de-raining task.}
  \label{fig:de-raining}
\end{figure}

\textbf{Single image de-raining.}
%Image de-raining and de-snowing~\cite{Zhang2017Image} can be posed as translating an input rainy or snowy image into a corresponding output clean image, but the background structure doesn't need to be changed.
We trained and tested our DRPAN model on single image de-raining task using the procedure as same as~\cite{Zhang2017Image}, and evaluated the results by both qualitative and quantitative metrics. Fig.~\ref{fig:de-raining} shows the qualitative results of our DRPAN with different sizes of discriminative region compared to ID-CGAN~\cite{Zhang2017Image}, and DRPAN outperforms ID-CGAN with not only more effective de-raining but also more vivid color and clear details. Tab.~\ref{tab:de-raining} reports the corresponding quantitative results evaluated by PSNR, SSIM, VIF, and RECO metrics, and the best results (in bold font) are achieved all by our DRPAN.

% single image de-raining

\begin{table}[!ht]\footnotesize
\centering
\caption{Quantitative comparison of our DRPAN (with different sizes of discriminative region) with ID-CGAN~\cite{Zhang2017Image} and PAN~\cite{Wang2017Perceptual} on image de-raining. DRPAN performs best (in bold font) evaluated by PSNR, SSIM, VIF, and RECO metrics}
  \begin{tabular}{p{0.1\columnwidth}|p{0.08\columnwidth}|p{0.12\columnwidth}|p{0.09\columnwidth}|p{0.14\columnwidth}|p{0.10\columnwidth}|p{0.10\columnwidth}|p{0.10\columnwidth}|p{0.10\columnwidth}}
    \hline
    Method  Metrics & L2+ CGAN & ID-CGAN\cite{Zhang2017Image} & PAN\cite{Wang2017Perceptual} & DRPAN (w/o mask) & DRPAN (128) & DRPAN (64) & DRPAN (32) & DRPAN (16)\\
    \hline\hline
    PSNR & 22.19 & 22.91 & 23.35 & 25.51 & 25.87 & 25.76 & 25.92 & \textbf{26.20}\\
    SSIM & 0.8083 & 0.8198 & 0.8303 & 0.8688 & 0.8714 & 0.8765 & \textbf{0.8788} & 0.8712\\
    VIF & 0.3640 & 0.3885 & 0.4050 & 0.4923 & 0.4818 & 0.4962 & \textbf{0.5001} & 0.4783\\
    RECO & -- & -- & -- & 0.9670 & 1.0770 & \textbf{1.1072} & 1.1067 & 1.0875\\
    \hline
  \end{tabular}
  \label{tab:de-raining}
\end{table}

\begin{figure}[!ht]
  \centering
  \includegraphics[width=.9\textwidth]{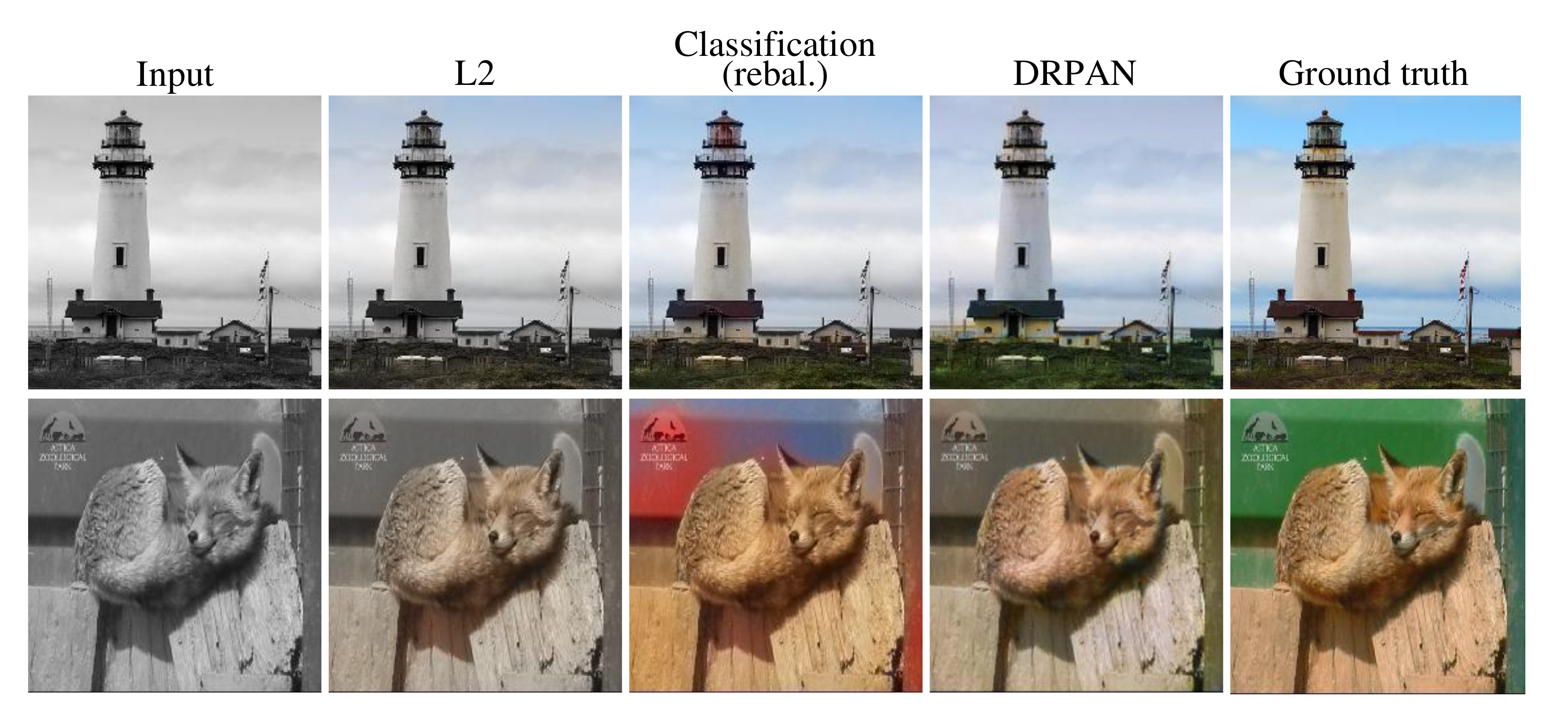}
  \caption{Example results of our DRPAN compared to L2 regression~\cite{Zhang2016Colorful} and Classification (rebal.)~\cite{Zhang2016Colorful} on image colorization task.}
  \label{fig:bw2color}
\end{figure}

\textbf{Bw to color.}
We trained our DRPAN model for image colorization task on ImageNet~\cite{deng2009imagenet}, and tested on ImageNet val dataset with an example shown in Fig.~\ref{fig:bw2color}. Our DRPAN can produce compelling colorization results compared with classification with class rebalancing~\cite{Zhang2016Colorful}. In addition, we run AMT evaluation for colorization(Tab.~\ref{tab:corlorization}). Our method fooled participants on $27.8\%$ which is competitive with the full method from~\cite{Zhang2016Colorful}.

\begin{table}[!ht]\footnotesize
\centering
\caption{AMT ``real vs fake'' test on corlorization}
  \begin{tabular}{p{0.35\textwidth}|p{0.4\textwidth}}
    \hline
    Method & \% Turkers labeled real\\
    \hline\hline
    L2 regression & 23.4\% \\
    Classification & \textbf{29.7\%} \\
    DRPAN & 27.8\% \\
    \hline
  \end{tabular}
  \label{tab:corlorization}
\end{table}

\subsection{Real to abstract translation}
We then implement our proposed DRPAN on two tasks of real to abstract translation which requires many-to-one abstraction ability.

\begin{figure}[!ht]
  \centering
  \includegraphics[width=\textwidth]{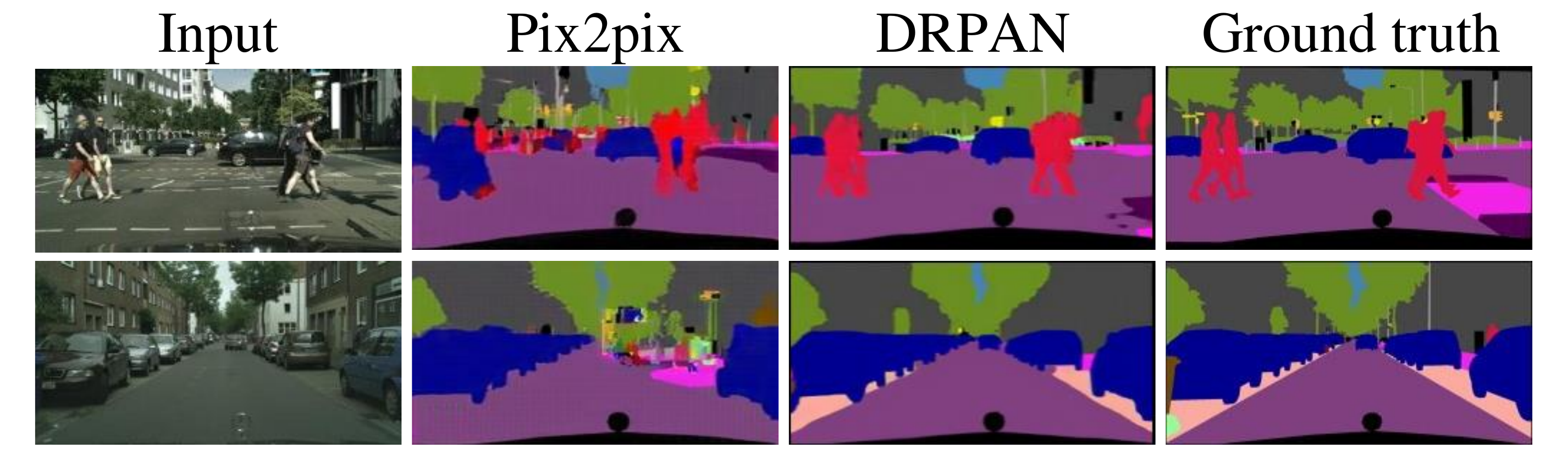}
  \caption{Example results of our DRPAN compared to Pix2pix~\cite{Isola2016Image} on real to semantic labels task.}
  \label{fig:seg}
\end{figure}

\textbf{Real to semantic labels.}
For real to semantic labels task, we tested our DRPAN model on two of the most used datasets: Cityscapes and facades. Fig.~\ref{fig:seg} shows the qualitative results of our DRPAN compared to Pix2pix~\cite{Isola2016Image} on Cityscapes dataset for translating real to semantic labels, and DRPAN can synthesize more realistic results that are closer to ground truth than Pix2pix, meanwhile, the quantitative results in Tab.~\ref{tab:seg} can also tell this in terms of per-pixel accuracy, per-class accuracy, and Class IOU.

\begin{figure}[!ht]
  \centering
  \includegraphics[width=\textwidth]{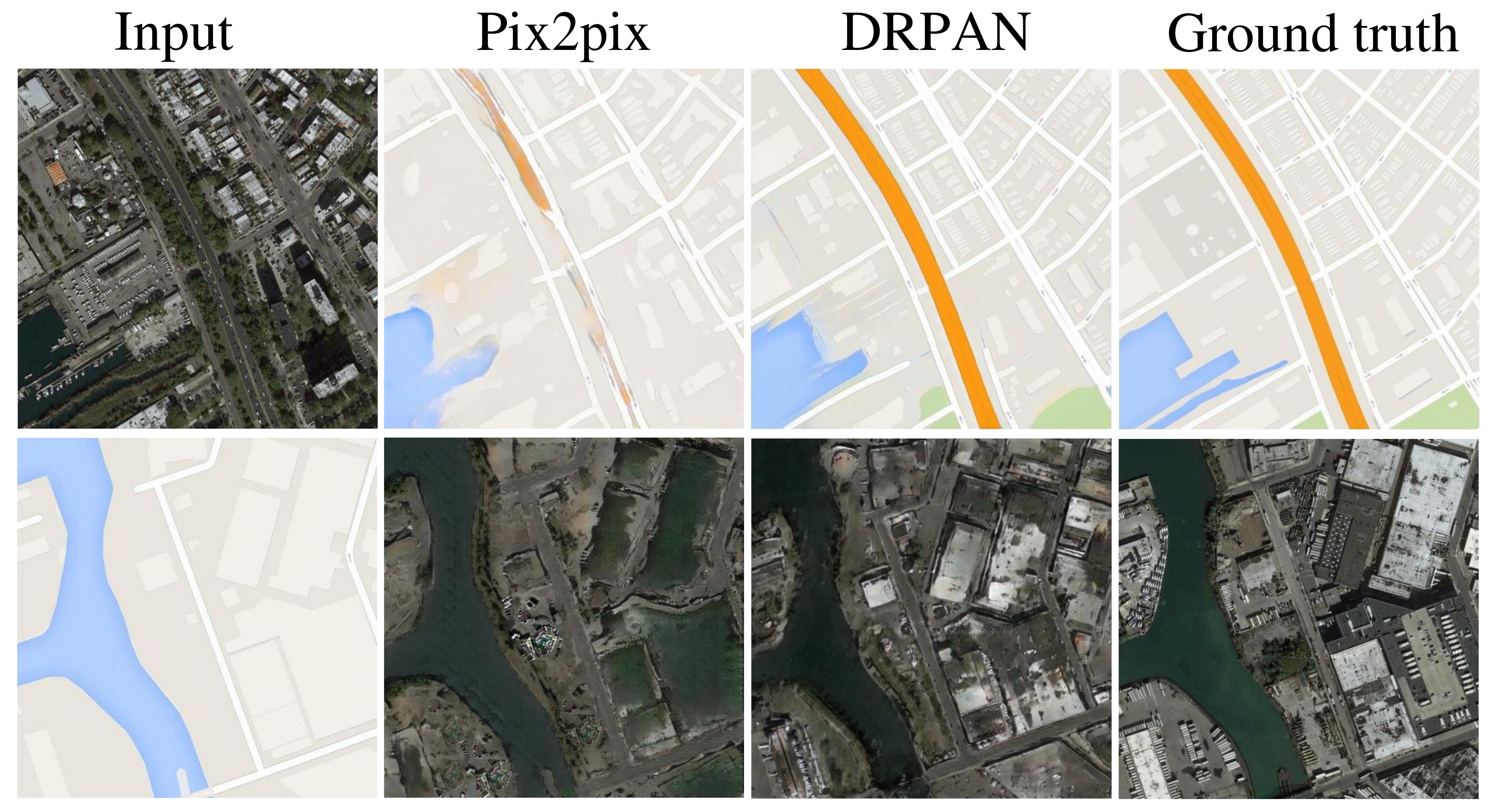}
  \caption{Example results of our DRPAN compared to Pix2pix~\cite{Isola2016Image} on aerial to maps (top) and maps to aerial (bottom) tasks.}
  \label{fig:maps}
\end{figure}

\textbf{Aerial to maps.}
We also applied our DRPAN on aerial photo to maps task, and the experiment was implemented using paired images with $512\times 512$ resolution~\cite{Isola2016Image}. The top row of Fig.~\ref{fig:maps} shows the qualitative results of our DRPAN compared to Pix2pix~\cite{Isola2016Image}, indicating that our DRPAN can correctly translate the motorway on aerial photo into the orange line on the map while Pix2pix can't.

\subsection{Abstract to real translation}
Besides, we also demonstrate our proposed DRPAN on several abstract to real tasks that can translate one to many: semantic labels to photo, maps to aerial, edge to real, and sketch to real.

\begin{figure}[!ht]
  \centering
  \includegraphics[width=.9\textwidth]{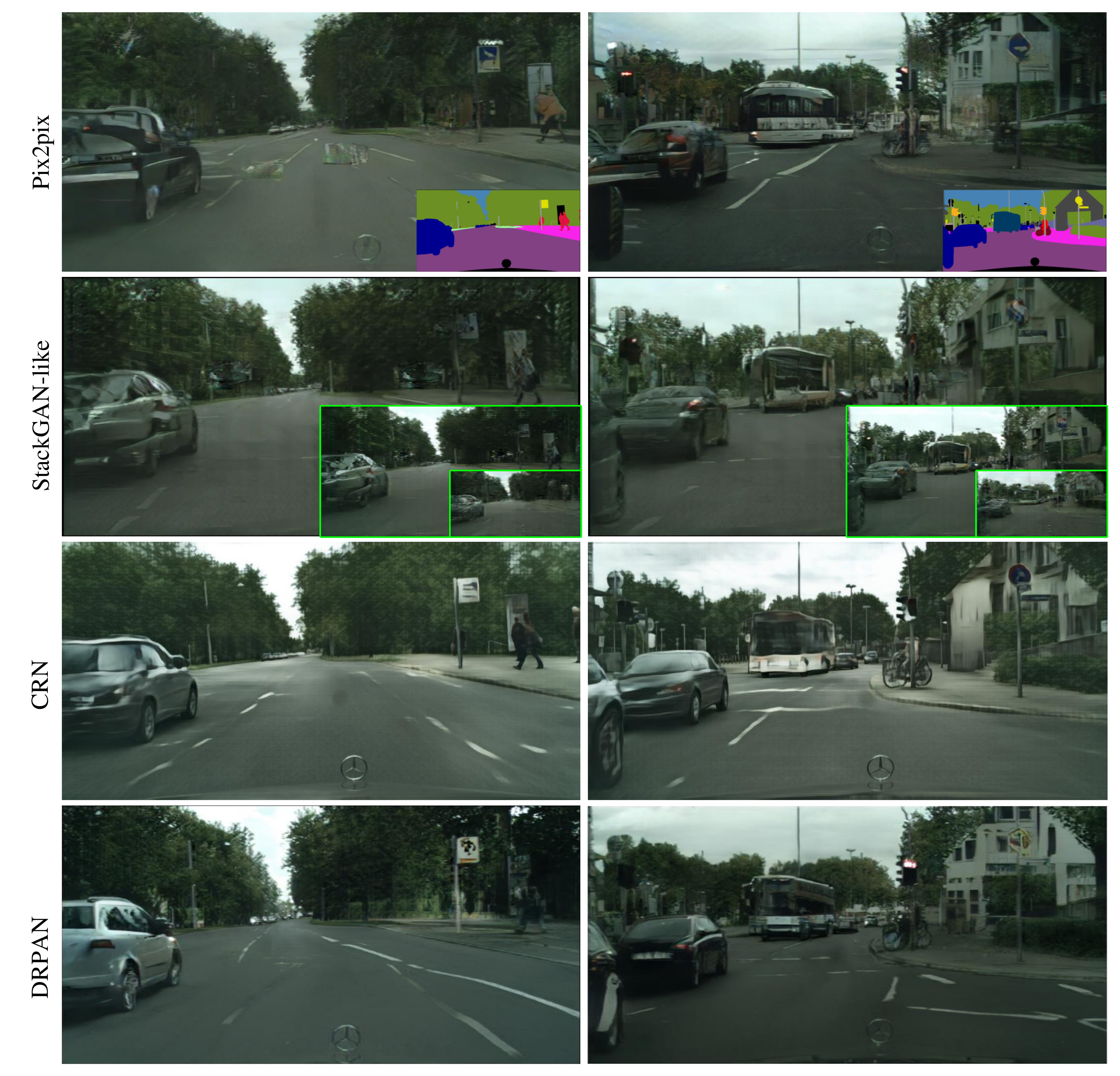}
  \caption{Example results of our DRPAN compared to Pix2pix~\cite{Isola2016Image} and CRN~\cite{Chen2017Photographic} on semantic labels to real task with $512 \times 512$ resolution.}
  \label{fig:cityscape-large}
\end{figure}

\textbf{Semantic labels to real.}
For semantic labels to real task, the translation model aims to synthesize real world images from semantic labels. CGAN based works fail to capture the details in the real world and suffer from deformation and blur problems. CNN based methods such as CRN can synthesize high-resolution but smooth rather than realistic results. Fig.~\ref{fig:cityscape-large} shows qualitative comparison of results, from which it can be seen that our DRPAN can synthesize the most realistic results with high-quality (more clear and less distorted while high resolution) compared to Pix2pix~\cite{Isola2016Image} and CRN~\cite{Chen2017Photographic}. 
%The inception score on Cityscapes and facades datasets in Tab.~\ref{tab:inceptionscore} also validates that our DRPAN performs better than Pix2pix.

The evaluation of GAN is still a challenging problem. Many works~\cite{Salimans2016Improved,Wang2016Generative,Zhang2016Colorful,Isola2016Image} used off-the-shelf classifiers as automatic measures of synthesized images. Tab.~\ref{tab:fcn8s} reports performance evaluation on segmentation of FCN-8s model, and our DRPAN exceeds Pix2pix~\cite{Isola2016Image} by $10\%$ on per-pixel accuracy and also achieves highest performance on per-class accuracy and Class IOU.

\begin{table}[!ht]
\begin{minipage}[t]{0.48\textwidth}
    \centering
    \caption{Quantitative comparison of our DRPAN with Pix2pix~\cite{Isola2016Image} on real to semantic labels task (Cityscapes dataset)}
    {\scriptsize
   \begin{tabular}{p{0.4\textwidth}|p{0.2\textwidth}|p{0.2\textwidth}|p{0.13\textwidth}}
    \hline
    Model &  Per-pixel acc. & Per-class acc. & Class IOU \\
    \hline\hline
    L1+U-Net~\cite{Isola2016Image} & 0.86 & 0.42 & 0.35\\
    \hline
    Pix2pix~\cite{Isola2016Image} & 0.83 & 0.36 & 0.29\\
    \hline
    DRPAN(w/o fake-mask) & \textbf{0.86} & \textbf{0.48} & \textbf{0.39}\\
    \hline
    DRPAN & \textbf{0.88} & \textbf{0.52} & \textbf{0.43}\\
    \hline
  \end{tabular}}
    \makeatletter\def\@captype{table}\makeatother
    \label{tab:seg}
\end{minipage}
\begin{minipage}[t]{0.48\textwidth}
    \centering
     \caption{Quantitative comparison of our DRPAN with other models on semantic labels to real task (Cityscapes dataset) by FCN-8s score}
    {\scriptsize
    \begin{tabular}{p{0.4\textwidth}|p{0.2\textwidth}|p{0.2\textwidth}|p{0.13\textwidth}}
    \hline
    Model & Per-pixel acc. & Per-class acc. & Class IOU\\
    \hline\hline
    %L1  & 0.44 & 0.14 & 0.10\\
    %\hline
    %CGAN  & 0.61 & 0.21 & 0.16\\
    %\hline
    %L1+GAN & 0.64 & 0.19 & 0.15\\
    %\hline
    L1+CGAN~\cite{Isola2016Image} & 0.63 & 0.21 & 0.16\\
    \hline
    CRN & 0.69 & 0.21 & \textbf{0.20}\\
    \hline
    DRPAN(w/o fake-mask) & \textbf{0.72} & \textbf{0.22} & 0.19\\
    \hline
    DRPAN & \textbf{0.73} & \textbf{0.24} & 0.19\\
    \hline
    Ground truth & 0.80 & 0.26 & 0.21\\
    \hline
    \end{tabular}}
    \label{tab:fcn8s}
   \end{minipage}
\end{table}

\begin{table}[!ht]
\begin{minipage}[t]{0.55\textwidth}
    \centering
    \caption{AMT real vs. fake results test on Cityscapes semantic labels to photo task}
    {\scriptsize
    \begin{tabular}{p{0.45\textwidth}|p{0.45\textwidth}}
    \hline
    Model & \% Turkers labeled real\\
    \hline\hline
    Pix2pix~\cite{Isola2016Image}  & 5.3\%\\
    \hline
    StackGAN-like~\cite{ZhangXLZHWM16} & 6.8\%\\
    \hline
    CRN~\cite{Chen2017Photographic} & 9.4\%\\
    \hline
    DRPAN(w/o fake-mask) & \textbf{14.3\%}\\
    \hline
    DRPAN & \textbf{18.2\%}\\
    \hline
     & \% Turkers labeled more realistic\\
    \hline\hline
    DRPAN vs. Pix2pix~\cite{Isola2016Image} & \textbf{91.2\%}\\
    \hline
    DRPAN vs. StackGAN-like & \textbf{84.6\%}\\
    \hline
    DRPAN vs. CRN~\cite{Chen2017Photographic} & \textbf{75.7\%}\\
    \hline
    \end{tabular}}
    \makeatletter\def\@captype{table}\makeatother
    \label{tab:AMTcityscape}
\end{minipage}
\begin{minipage}[t]{0.45\textwidth}
    \centering
    \caption{AMT real vs. fake results test on maps to aerial task}
    {\scriptsize
    \begin{tabular}{p{0.45\textwidth}|p{0.45\textwidth}}
    \hline
    Model & \% Turkers labeled real\\
    \hline\hline
    Pix2pix~\cite{Isola2016Image}  & 25.2\%\\
    \hline
    DRPAN(w/o fake-mask) & 31.7\%\\
    \hline
    DRPAN & \textbf{33.4\%}\\
    \hline
    \end{tabular}}
    \makeatletter\def\@captype{table}\makeatother
    \label{tab:AMTmaps2aerial}
   \end{minipage}
\end{table}

\textbf{Maps to aerial.}
As opposed to aerial to maps task, we also tested our DRPAN on maps to aerial task, and the qualitative results are shown in the bottom row of Fig.~\ref{fig:maps}, which clearly demonstrates that our DRPAN can synthesize higher quality aerial photos than Pix2pix~\cite{Isola2016Image}.

\textbf{Human perceptual validation.}
We assess the performance of abstract to real on semantic labels to photo and maps to aerial by AMT. For fake against real study, we followed the perceptual study protocol from~\cite{Isola2016Image}, and collected data of each algorithm from $30$ participants. Each participant has $1000ms$ to look one sample. We also compared how realistic the synthesized images between different algorithms. Tab.~\ref{tab:AMTcityscape} illustrates that images synthesized by DRPAN are ranked more realistic than state-of-the-arts (DRPAN $18.2\%>$ CRN $9.4\%>$ StackGAN-like $6.8\%>$ Pix2pix $5.3\%$), moreover, compared to Pix2pix~\cite{Isola2016Image}, StackGAN-like~\cite{ZhangXLZHWM16} and CRN~\cite{Chen2017Photographic}, images synthesized by DRPAN are ranked more realistic by $91.2\%$, $84.6\%$ and $75.7\%$ respectively. Tab.~\ref{tab:AMTmaps2aerial} reports the comparison on maps to aerial task and our DRPAN fooled participants on $39.0\%$ over $18.7\%$ of Pix2pix and $26.8\%$ of CycleGAN~\cite{ZhuPIE17} respectively.

\begin{figure}[!ht]
  \centering
  \includegraphics[width=\textwidth]{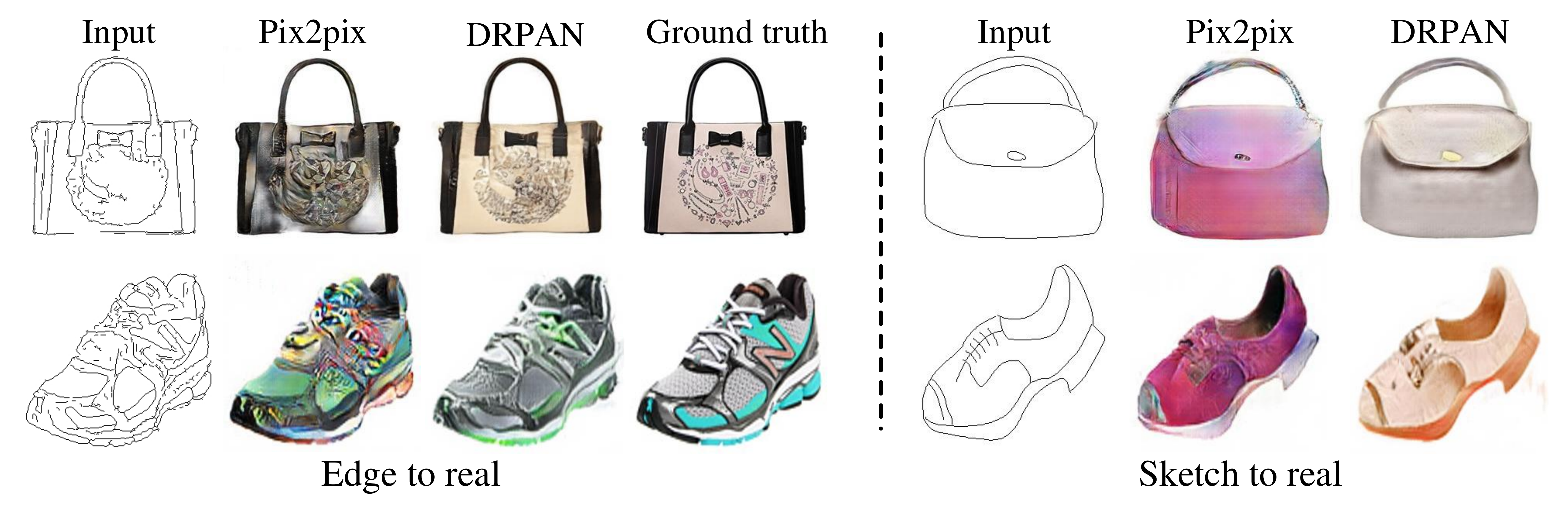}
  \caption{Example results of our DRPAN compared to Pix2pix~\cite{Isola2016Image} on edge to real (left) and sketch to real (right) tasks.}
  \label{fig:edge}
\end{figure}

\textbf{Edges to real and sketch to real.}
For the edge to real and sketch to real tasks, previous works often encounter with two problems~\cite{Isola2016Image}: one is that it's easy to generate artifacts and artificial color distribution in regions when the input such as edge is sparse; the other is that it's difficult to deal with unusual inputs like sketch. We tested our DRPAN model on UT Zappos50k dataset~\cite{fine-grained} and edge to handbag dataset~\cite{zhu2016generative}. Fig.~\ref{fig:edge} shows that our model can also handle these two problems well.

\section{Conclusions}
We propose Discriminative Region Proposal Adversarial Networks (DRPAN) towards high-resolution and photo-reality image-to-image translation. Human perceptual studies and automatic quantitative measures validate the performance of our proposed DRPAN against the state-of-the-arts for synthesizing high-quality results. We hope it can be explored for discriminative feature learning and other computer vision tasks in the future.

\subsection*{Acknowledgments}
This work was supported by the National Natural Science Foundation of China under Grants 61771440 and 41776113, and Qingdao Municipal Science and Technology Program under Grant 17-1-1-5-jch.
%
% ---- Bibliography ----
%
% BibTeX users should specify bibliography style 'splncs04'.
% References will then be sorted and formatted in the correct style.
%
\bibliographystyle{splncs04}
\bibliography{mybib}
%
% \begin{thebibliography}{8}
% \bibitem{ref_article1}
% Author, F.: Article title. Journal \textbf{2}(5), 99--110 (2016)

% \bibitem{ref_lncs1}
% Author, F., Author, S.: Title of a proceedings paper. In: Editor,
% F., Editor, S. (eds.) CONFERENCE 2016, LNCS, vol. 9999, pp. 1--13.
% Springer, Heidelberg (2016). \doi{10.10007/1234567890}

% \bibitem{ref_book1}
% Author, F., Author, S., Author, T.: Book title. 2nd edn. Publisher,
% Location (1999)

% \bibitem{ref_proc1}
% Author, A.-B.: Contribution title. In: 9th International Proceedings
% on Proceedings, pp. 1--2. Publisher, Location (2010)

% \bibitem{ref_url1}
% LNCS Homepage, \url{http://www.springer.com/lncs}. Last accessed 4
% Oct 2017
% \end{thebibliography}
\end{document}